\documentclass{article}

\usepackage{PRIMEarxiv}

\usepackage[utf8]{inputenc} 
\usepackage[T1]{fontenc}    
\usepackage{hyperref}       
\usepackage{url}            
\usepackage{booktabs}       
\usepackage{amsfonts}       
\usepackage{nicefrac}       
\usepackage{microtype}      
\usepackage{lipsum}
\usepackage{media9}
\usepackage{fancyhdr}       
\usepackage{graphicx}       
\graphicspath{{media/}}     

\title{DiTPainter: Efficient Video Inpainting with Diffusion Transformers
}

\author{
  Xian Wu, Chang Liu \\
  ByteDance \\
}

\begin{document}
\maketitle

\begin{abstract}
Many existing video inpainting algorithms utilize optical flows to construct the corresponding maps and then propagate pixels from adjacent frames to missing areas by mapping. Despite the effectiveness of the propagation mechanism, they might encounter blurry and inconsistencies when dealing with inaccurate optical flows or large masks. Recently, Diffusion Transformer (DiT) has emerged as a revolutionary technique for video generation tasks. However, pretrained DiT models for video generation all contain a large amount of parameters, which makes it very time consuming to apply to video inpainting tasks. In this paper, we present DiTPainter, an end-to-end video inpainting model based on Diffusion Transformer (DiT). DiTPainter uses an efficient transformer network designed for video inpainting, which is trained from scratch instead of initializing from any large pretrained models. DiTPainter can address videos with arbitrary lengths and can be applied to video decaptioning and video completion tasks with an acceptable time cost. Experiments show that DiTPainter outperforms existing video inpainting algorithms with higher quality and better spatial-temporal consistency.
\end{abstract}


\section{Introduction}
Video inpainting aims to fill in plausible pixels in the missing video area. It has a wide range of applications in the field of video editing, such as object removal, video completion, and video decaptioning. Video inpainting is challenging because it is supposed to generate video contents which are visually realistic as well as spatial-temporal consistent with surroundings. In addition to that, one needs to inpaint pixels for all frames in the video, which is usually time-consuming and hinders practical application.

Many existing works \cite{zhang2022flow,liCvpr22vInpainting,zhou2023propainter} use propagation-based algorithms to address the video inpainting task. They utilize optical flows to find the corresponding contents from adjacent frames and then propagate the consistent information to the missing areas. Propagation-based methods can achieve surprising results with complete corresponding contents. However, these methods suffer from inaccurate optical flows or unknown corresponding video contents, which may lead to temporal inconsistencies. When dealing with large masks, these methods usually produce results with blurry and artifacts, as it is hard for them to synthesize new content of high quality without enough guidance information.


Diffusion models have already demonstrated their powerful capabilities and achieved significant progress in the field of visual generation. More recently, Peebles and Xie \cite{Peebles_2023_ICCV} present Diffusion Transformer (DiT), which combines diffusion models with a transformer-based architecture. DiT has emerged as a revolutionary technique for video generation tasks, because it complies with the scaling law and excels at maintaining temporal consistency. Several DiT-based video generation models \cite{yang2024cogvideox,polyak2025moviegencastmedia,lin2024open} have achieved impressive results, but they all contain a large number of parameters. It is very time-consuming to use these large pretrained models for downstream applications. For video inpainting applications, such as video decaptioning and object removal, users are often sensitive to time consumption, making it difficult to apply large pretrained models directly.

To address above concerns, we present DiTPainter in this paper, an efficient video inpainting model based on Diffusion Transformer (DiT). Unlike most video editing methods recently, which use a large pretrained model and finetune from it for downstream applications, DiTPainter adopts a self-designed small transformer-based network which is trained from scratch. This small DiT model significantly saves computational cost for inference and gpu resources for training. There is no text description input into DiTPainter, which makes it more convenient for practical applications and without the need for visual-text cross attentions. DiTPainter adopts a 3D VAE to encode video frames into the latent space and downsample both spatial and temporal dimensions. It also utilizes Flow Matching \cite{lipman2023flowmatchinggenerativemodeling} to reduce inference steps for efficiency and achieves satisfactory performance even in 4 or 8 steps. To deal with long videos, we employ MultiDiffusion to DiTPainter \cite{bartal2023multidiffusionfusingdiffusionpaths} for the temporal consistency of transition frames, making it convenient to apply to video decaptioning and video completion tasks. Experiments show that DiTPainter can produce competitive results with an acceptable time cost, compared to existing video inpainting algorithms.

\begin{figure}[h]
    \centering
    \includegraphics[width=1.0\textwidth]{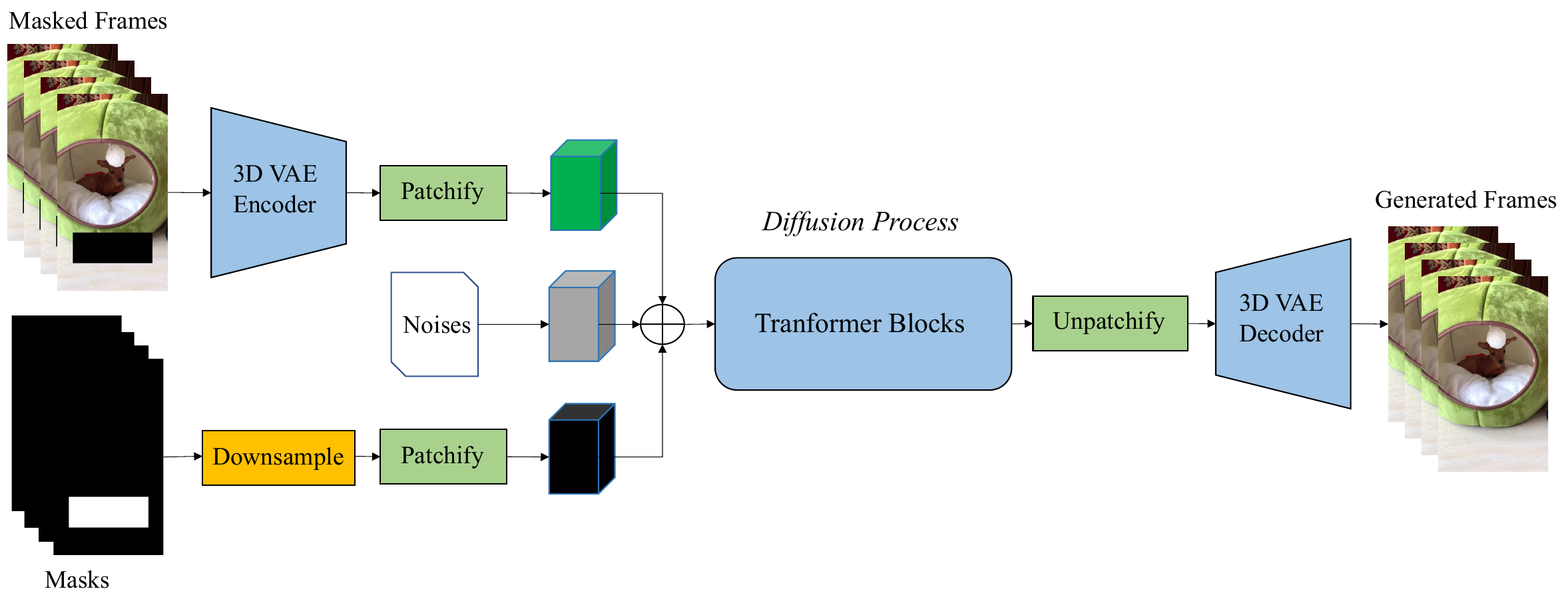}
    \caption{Pipeline of our method. Masked frames are first encoded into 3D latents and corresponding masks are downsampled to the same size. We patchify video latents, masks along with random noises and add them together as a sequence of tokens. After the diffusion process conducted through several transformer blocks, we can decode tokens into video frames as our final results.}
    \label{fig:pipeline}
\end{figure}

\section{Method}

Given a masked video sequence $Y \in \mathbb{R}^{H\times W\times N \times 3}$ with $N$ frames, along with corresponding masks $M \in \mathbb{R}^{H\times W\times N \times 1}$, video inpainting aims to generate plausible visual contents in the masked area which should be consistent and coherent with surrounding pixels. At first, DiTPainter uses 3D VAE to encode $Y$ into video latents $y$ and downsample its time-space dimensions, as $y \in \mathbb{R}^{h\times w\times n \times c}$. Then latents $y$, downsampled masks $m$ and noises with the same size are all fed into a transformer-based network. After several denoising steps, video latents output by Diffusion Transformers are decoded into the final results $X \in \mathbb{R}^{H\times W\times N \times 3}$. For varying lengths of video sequences, we utilize MultiDiffusion in the temporal axis for the consistency of transition frames between video clips. Figure \ref{fig:pipeline} illustrates the inference pipeline of our method.

\textbf{3D VAE.} Similar to recent video generation frameworks, we use a 3D VAE to encode masked frames $Y$ into the latent space for video compression. We adopt the pretrained model of WF-VAE \cite{li2024wf} for convenience. WF-VAE utilizes 3D and 2D wavelet transforms in the frequency domain and leverages multi-level features by a pyramid structure. It achieves high reconstruction quality with fast encoding and decoding speed, which can benefit efficient video inpainting.

Given masked video frames $Y \in \mathbb{R}^{H\times W\times N\times 3}$, the encoder compresses them into low-dimensional latents $y \in \mathbb{R}^{h\times w\times n\times c}$, where $h=H/8, w=W/8, n=(N-1)/4+1$ and $c=8$. Corresponding masks $M \in \mathbb{R}^{H\times W\times N\times 1}$ are also downsampled to $m$ with the same size $h \times w \times n$. We find in experiments that missing temporal information of masks may lead to flickers in inpainting results. Therefore, we recover the reduced temporal dimension of masks through the channel dimension for completeness, as $m \in \mathbb{R}^{h\times w\times n\times 4}$. In the decoding stage, the denoised latents are decoded into the original resolution as inpainting frames $X \in \mathbb{R}^{H\times W\times N\times 3}$.

\textbf{Diffusion Transformer.} For masked video latents $y$, downsampled masks $m$ and random noises, we first patchify them of the same $2 \times 2 \times 1$ spatial-temporal size and project all 3D patches into the same embedding dimension. The three embeddings are then flattened into 1D sequences and added together as input tokens for transformer blocks. Therefore, the total length of the input tokens is $w \times h \times n/4$.

\begin{figure}
    \centering
    \includegraphics[width=0.3\textwidth]{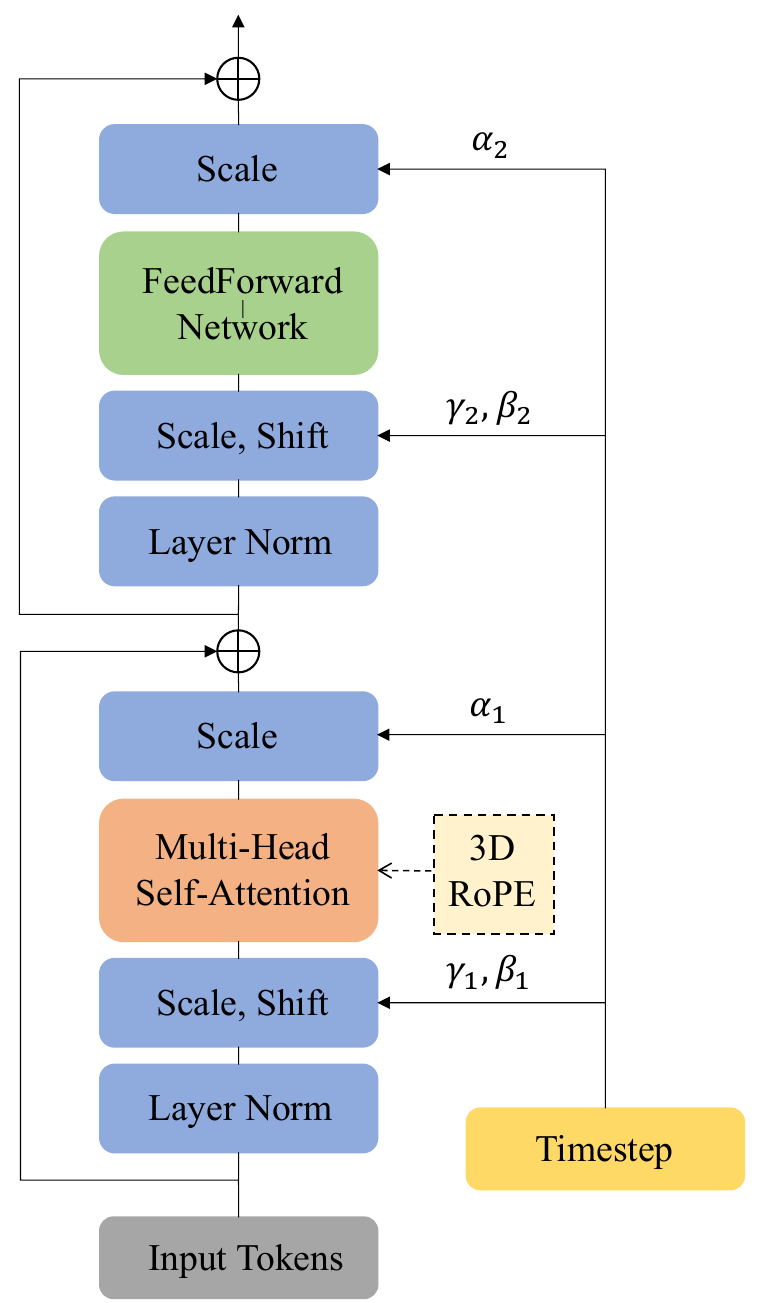}
    \caption{Structure of our transformer block.}
    \label{fig:block}
\end{figure}

We adopt a pre-norm transformer block structure primarily comprising a multi-head self-attention and a feedforward network. Our transformer block excludes the cross-attention as there is no text condition. We regress two sets of scale and shift parameters from timesteps through \textit{adaLN-Zero block}, and then inject them into the self-attention and the FFN separately. Following recent text-to-video generation models \cite{yang2024cogvideox,lin2024open,isobe2025amdhummingbirdefficienttexttovideomodel}, we also utilize 3D Full Attention and 3D RoPE \cite{su2023roformerenhancedtransformerrotary} to enhance video smoothness and quality. After the final transformer block, we project each token into a $2 \times 2 \times 2c$ tensor through a linear layer and unpatchify the 1D sequence back into the original 3D size. Figure \ref{fig:block} illustrates the structure of our transformer block.

Our DiT model consists of 24 transformer blocks. Each block uses 16 attention heads with 72 hidden dimensions. The total number of parameters in our DiT model is 0.4B, which is much less than most text-to-video generation models. Experiments show that our simple and small network can achieve competitive results in video inpainting tasks, although it is not based on any pretrained large model.

\textbf{Flow Matching.} Since Flow Matching demonstrates its ability to generate high-quality images and videos through few denoising steps, we utilize it as our diffusion scheduler for efficiency. During the training stage, given encoded latents $x_1$ of a source video, a random noise $x_0 \sim \mathcal{N}(0,I)$ following Gaussian distribution, 
and a timestep $t \in [0,1]$, $x_1$ is noised by $x_0$ using a linear interpolation as follows,
\begin{equation}
x_t=tx_1+(1-t)x_0.
\end{equation}
Our DiT model is then optimized to predict the velocity, namely, 
\begin{equation}
v_t = \frac{dx_t}{dt}=x_1-x_0.
\end{equation}
Thus, the loss function is the mean squared error between the model prediction and the ground truth velocity $v_t$, expressed as
\begin{equation}
\mathcal{L}=\mathbb{E}_{x_0,x_1,y,m,t}\|u(x_t,y,m;\theta)-v_t\|^2,
\end{equation}
where $\theta$ denotes the model parameters to be optimized and $u(x_t,y,m;\theta)$ is the model prediction. Following Stable Diffusion 3 \cite{esser2024scalingrectifiedflowtransformers}, the logit normal distribution is applied for $t$ to give more weight to intermediate timesteps.

In the inference stage, we first sample a random noise $x_0 \sim \mathcal{N}(0,I)$. Then we use the first-order Euler ODE solver to compute $x_t$ by integrating the model prediction as the velocity. Owing to Flow Matching, DiTPainter can generate plausible inpainting results even in 4 inference steps, as shown in experiments.

\textbf{Temporal MultiDiffusion.} To address longer videos with more frames than training, we apply MultiDiffusion to the temporal axis, making it effective for our model to inpaint videos of arbitrary length with temporal consistency. Since the denoising process is performed in the latent space, we consider the video latents $x$ for convenience. 

Given a video with $N'$ frames where $N' > N$, the length of its encoded latents $x'$ is $n'=(N'-1)/4+1$. We segment the latents $x'$ into several overlapping clips by a sliding window with a length of $n$ and a stride of $s$. This process partitions $x'$ into latent clips $\{x^k\}^r_{k=1}$, where $r=\lceil (n'-n)/s \rceil +1$ is the total number of clips. The denoising step is performed on each latent clip and we denote the $k$-th clip as $x^k_t$ at the timestep $t$.

For the $i$-th latent of the temporal index, denoted as $x'[i]$, we can find the set of clips $\mathcal{S}(i)=\{x^k|x'[i]\in x^k\}$ that contain this latent. For each clip $x^k$ in $\mathcal{S}(i)$, we denote the corresponding latent as $x^k[j]$ which is mapped from $x'[i]$. After the timestep $t$ denoising process performed on the clips, we update the value of $x'_t[i]$ by averaging all the corresponding latents:

\begin{equation}
x'_t[i]=\frac{1}{\|\mathcal{S}(i)\|}\sum_{x^k\in\mathcal{S}(i)}x^k_t[j].
\end{equation}

We update the values of all latents using the above formulation and then map them to the corresponding clips. Subsequently, the next denoising step is performed on the basis of the mapping values. Since overlapping clips share the same latent space, they are able to maintain temporal consistency at transition frames.

\section{Experiments}

\textbf{Training details.} We employ a two-stage coarse-to-fine strategy to train DiTPainter, since we find that convergence is difficult to achieve by training the model directly on high-resolution videos. At the first stage, we train the DiT model on 240p videos to capture spatial and temporal consistency in a coarse manner. At the second stage, we continue to train the model on 720p videos to enhance fine details for the high quality. DiTPainter is trained for 500k iterations at the first stage and 200k iterations at the second stage. The batch size of training is 16 and the length of video frames is 65. We use the AdamW optimizer with a constant learning rate of 1e-5. To simulate masks used in video decaptioning and video completion tasks, during training, we generate stationary and moving masks in a random pattern following ProPainter \cite{zhou2023propainter}.

\begin{figure}
    \centering
    \includegraphics[width=0.5\textwidth]{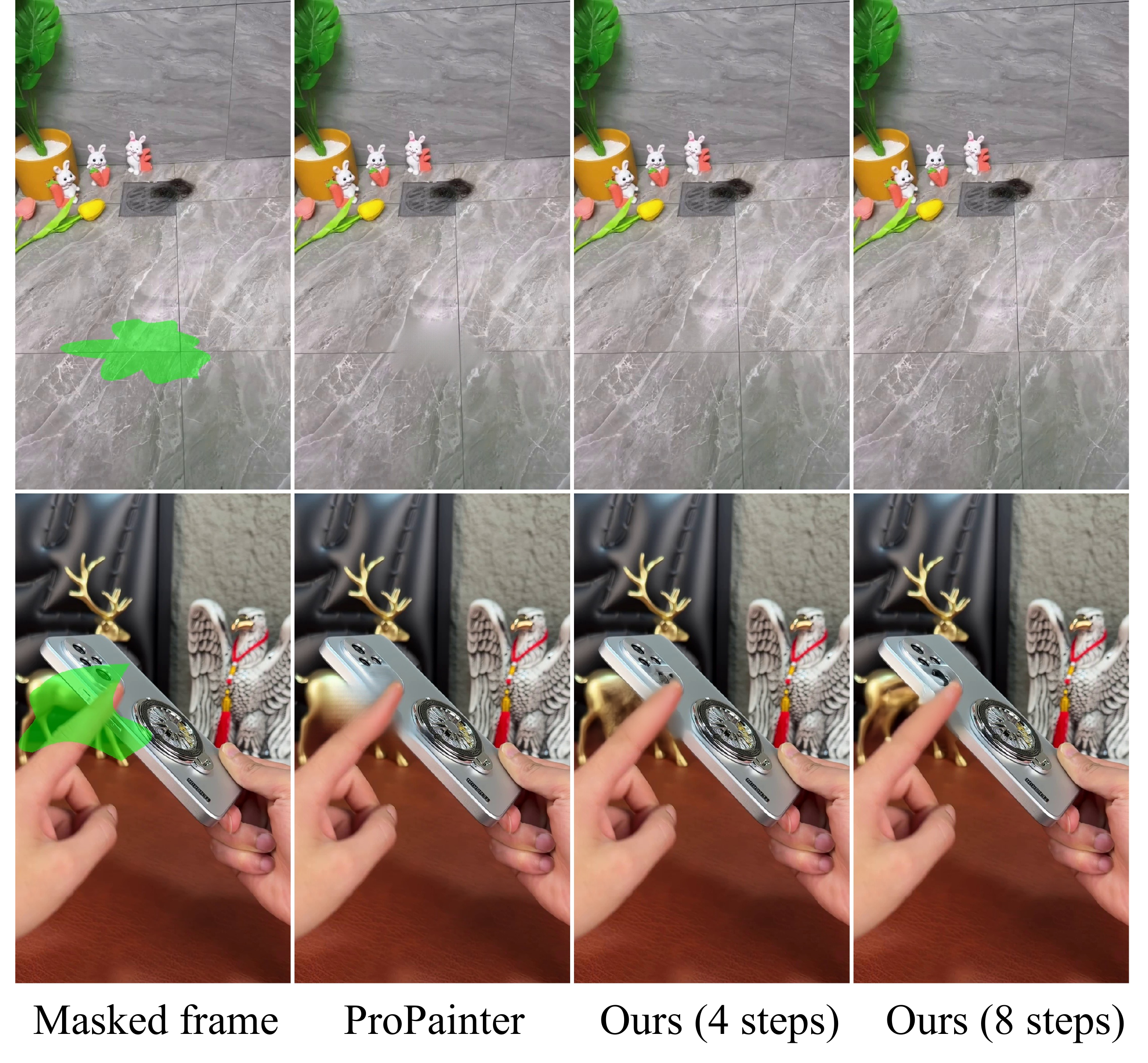}
    \caption{Results of video completion by ProPainter and our method.}
    \label{fig:completion}
\end{figure}

\textbf{Quantitative comparison.} We collect 50 short videos with the resolution of 720p as a test set. We generate masks by the same pattern of training and employ models to complete masked videos to calculate quantitative scores. Video frames are resized to $432 \times 240$ for evaluation. The quantitative scores in Table~\ref{tab:table} show that our method surpasses ProPainter, the state-of-the-art video inpainting algorithm. Due to Flow Matching, our method can address video inpainting even in 4 or 8 inference steps. Figure \ref{fig:completion} visualize some results of video completion in the test set.

\begin{table}
 \caption{Quantitative comparison on video completion.}
  \centering
  \begin{tabular}{llll}
    \toprule
    \cmidrule(r){1-2}
             & PSNR$\uparrow$ & SSIM$\uparrow$ & VFID$\downarrow$ \\
    \midrule
    ProPainter & 34.46  & 0.9834 & 0.069     \\
    Ours(4 steps) & \textbf{34.86} & \textbf{0.9844} & 0.056   \\
    Ours(8 steps) & 34.60 & 0.9843 & \textbf{0.051}  \\
    \bottomrule
  \end{tabular}
  \label{tab:table}
\end{table}

\begin{figure}[h]
    \centering
    \includegraphics[width=1.05\textwidth]{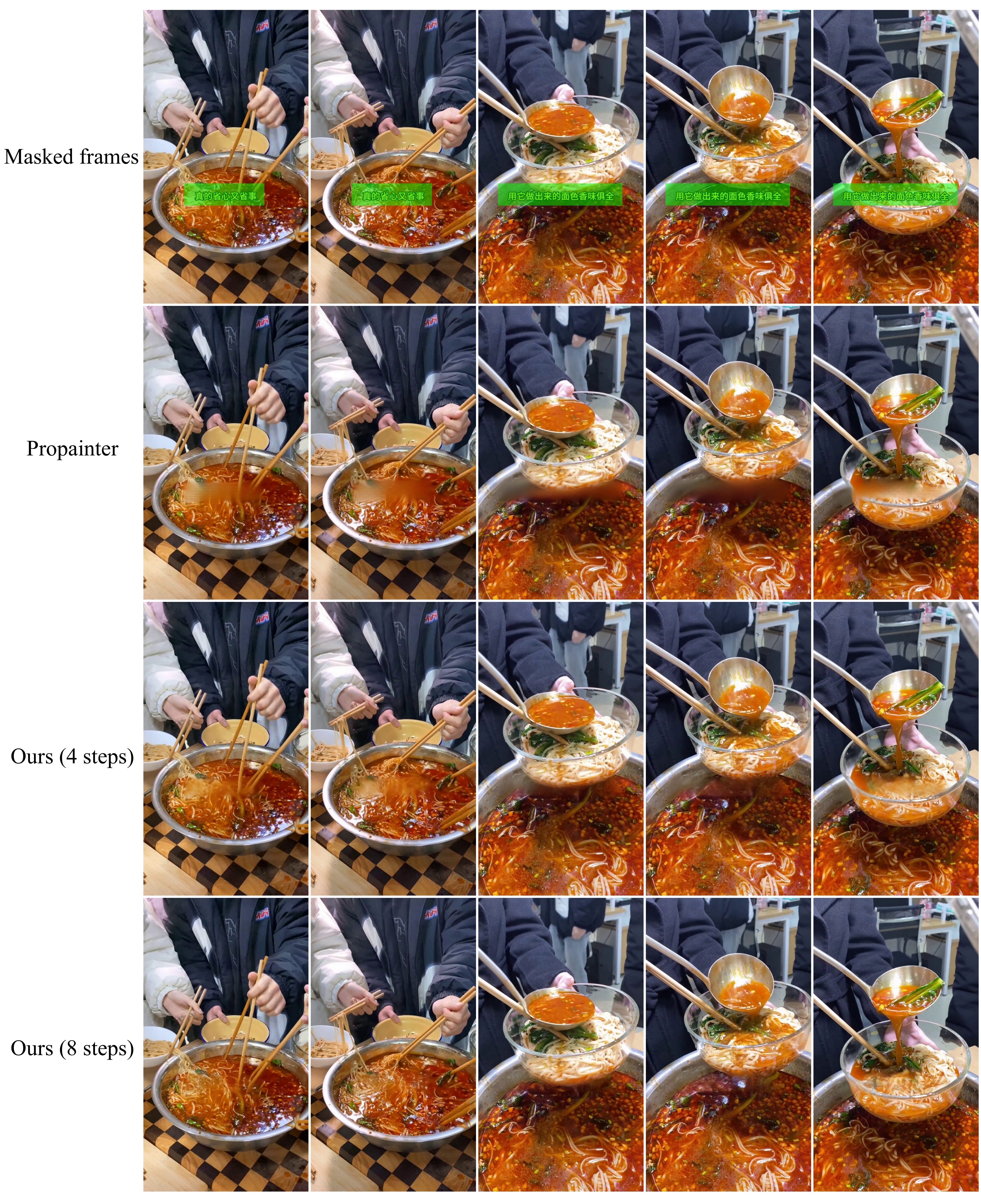}
    \caption{Qualitative comparison of video decaptioning between ProPainter and our method.}
    \label{fig:comp1}
\end{figure}

\begin{figure}[h]
    \centering
    \includegraphics[width=1.05\textwidth]{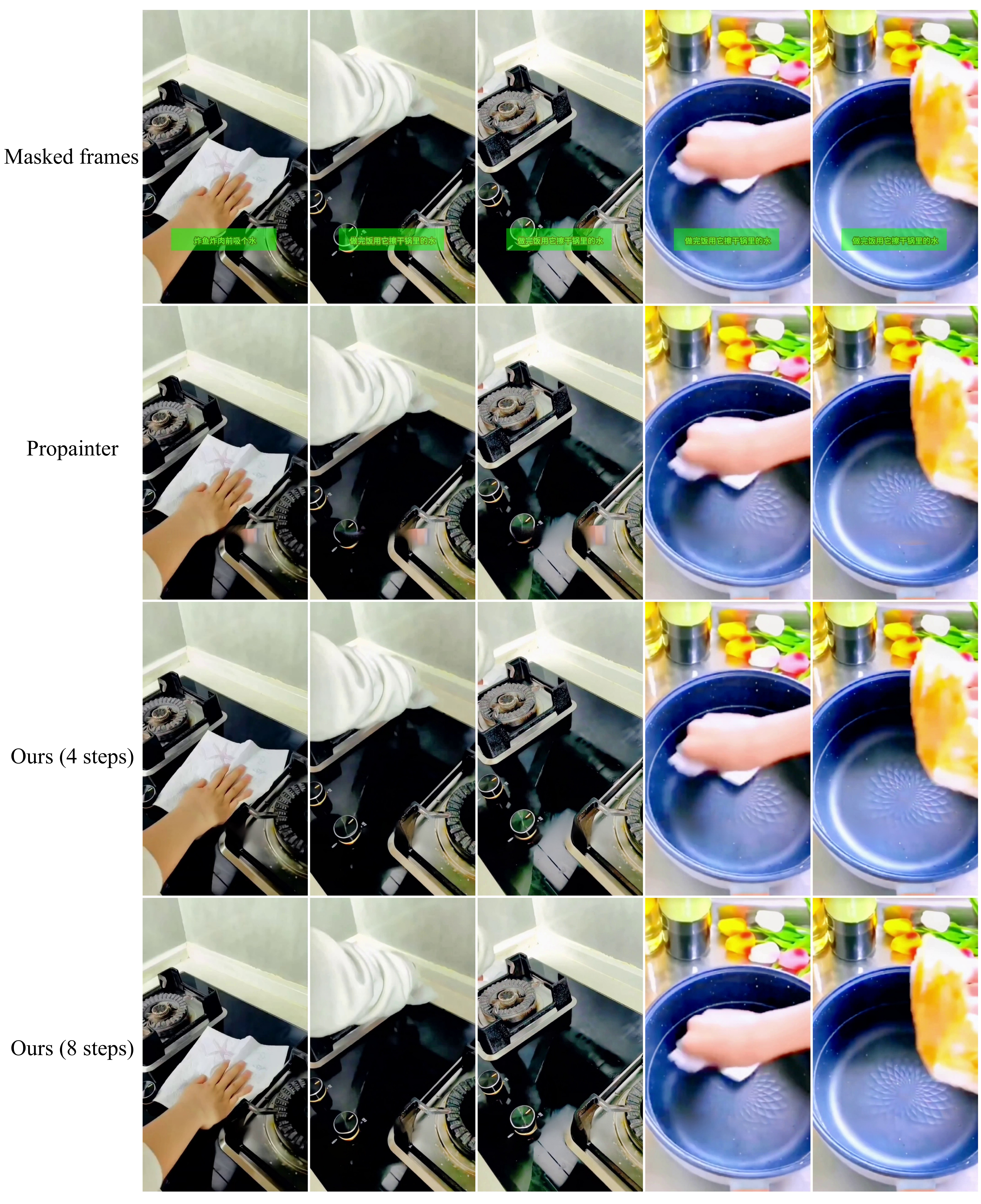}
    \caption{Qualitative comparison of video decaptioning between ProPainter and our method.}
    \label{fig:comp2}
\end{figure}

\textbf{Qualitative comparison.} It is convenient and efficient to apply our method to video editing applications, such as video decaptioning. Figures \ref{fig:comp1} and \ref{fig:comp2} show the results of video decaptioning using Propainter and DiTPainter. Propainter may cause blurry and artifacts while estimated optical flows are not accurate or there is no corresponding pixel for propagation. In contrast, our method can achieve high-quality results with spatial-temporal consistency and realistic textures.

\section{Conclusion}
In this paper, we present DiTPainter in this paper, a state-of-the-art video inpainting model based on Diffusion Transformer (DiT). Instead of relying on a large pretrained model, we carefully design a small transformer-based network and train it from scratch. This small DiT model significantly saves computational cost for efficient inference and training. DiTPainter adopts WF-VAE to encode video frames into the 3D latent space with downsampling. DiTPainter utilizes Flow Matching to generate plausible videos even in 4 or 8 inference steps. To deal with longer videos, DiTPainter uses MultiDiffusion to obtain temporal consistency of transition frames. DiTPainter can be applied to several video inpainting tasks with an acceptable time cost, such as video decaptioning and video completion. Qualitative comparisons show that DiTPainter outperforms the existing video inpainting algorithm with higher visual quality and better temporal consistency.

\bibliographystyle{unsrt}  
\bibliography{references}

\end{document}